\documentclass[11pt,a4paper]{article}
\usepackage[hyperref]{emnlp2020}
\usepackage{times}
\usepackage{latexsym}
\usepackage{graphicx}
\usepackage{tabularx}
\usepackage{wrapfig}
\usepackage{amsmath}
\usepackage{natbib}
\usepackage[T1]{fontenc}
\usepackage{babel}

\usepackage{microtype}

\aclfinalcopy % Uncomment this line for the final submission
\newcommand*\samethanks[1][\value{footnote}]{\footnotemark[#1]}

\title{Instructions for EMNLP 2020 Proceedings}

\author{Xiaoyu Chen\thanks{equal contribution} \\
  Dept. of Computer Science \\
  Stanford University \\
  \texttt{codchen@stanford.edu} \\\And
  Rohan Badlani\samethanks \\
  Dept. of Computer Science \\
  Stanford University \\
  \texttt{rohan.badlani@cs.stanford.edu} \\}

\title{Relation Extraction with Contextualized Relation Embedding (CRE)}

\begin{document}
\maketitle
\begin{abstract}
    Relation extraction (RE) is the task of identifying relation instance(s) between two entities given a corpus whereas Knowledge base (KB) modeling is the task of representing a knowledge base, in terms of relations between entities. This paper proposes an architecture for the relation extraction (RE) task that integrates semantic information with knowledge base (KB) 
    modeling in a novel manner. Existing approaches for relation extraction either don't utilize knowledge base modelling or use separately trained KB models for the RE task. We present a model architecture that internalizes KB modeling in 
    relation extraction. This model applies a novel approach to encode sentences into contextualized relation embeddings (CRE), which can then be used together with parameterized entity embeddings to score relation instances. The proposed CRE model achieves state of the art performance on datasets derived from The New York Times Annotated Corpus\footnote{\url{https://catalog.ldc.upenn.edu/LDC2008T19}} and FreeBase\footnote{\url{https://developers.google.com/freebase}}. 
    The source code has been made available\footnote{\url{https://github.com/codchen/CRE}}.
\end{abstract}

\section{Introduction}
Relation extraction (RE) is a sub-task under the broad category of information extraction (IE) that aims to identify 
relationship(s) between named entities based on textual information (corpus). The groundtruth relationship(s) between an 
entity pair can be either internal, if each sentence in the corpus is explicitly labeled, or external, in the form of relation instances in a standalone 
knowledge base (KB). Such knowledge bases, like Freebase \citet{freebase}, are collaboratively edited by human beings and thus offer high 
informational fidelity.

In this paper, we focus on the problem of predicting the collection of relations for a new entity pair based on the usage of that entity pair in its respective collection of sentences. To this end, we assume the general construction that given a dataset with a set of named entity pairs and each entity pair's respective 
collection of sentences as input along with entity pair’s respective collection of relations as output, our goal is to train a model with the objective of predicting collection of \textit{relevant} relation(s) for a new entity pair based on its contextual usage. 

Distant supervision \citet{mintz-2009} is the most popular approach for this problem, that has achieved its success by leveraging knowledge base information for relation extraction tasks. However, this restricts the usage of semantic information present in knowledge bases, since the distant supervision work mainly incorporates the knowledge base information as labels instead of treating it as a graph and thereby losing the dense relationships between different entities. Knowledge base modelling is an independent area of research and there has been some recent work on utilizing knowledge base modeling and incorporating internal structural information from them to the relation extraction task. \citet{transe}, \citet{complex}. 

\begin{figure*}
    \includegraphics[width=\textwidth]{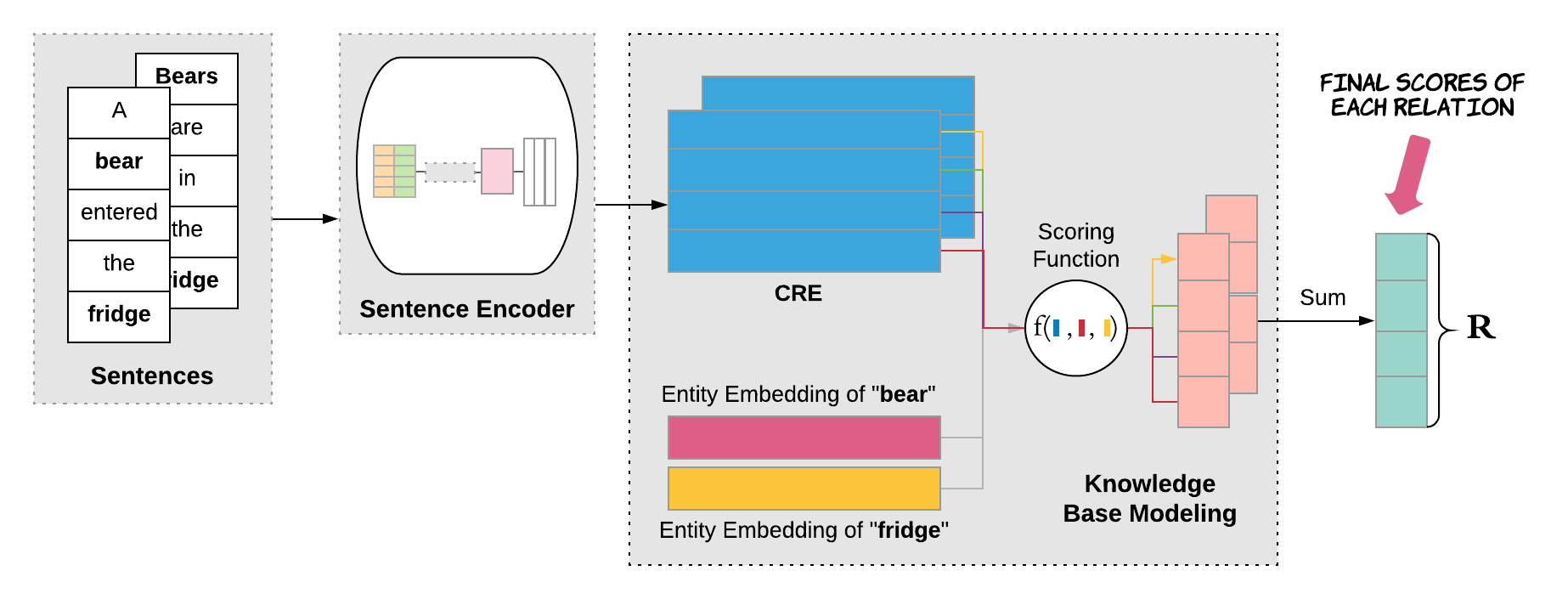}
    \caption{Contextualized Relation Embedding (CRE) Model Architecture}\label{fig:model}
    \medskip
    \small
    The details of "Sentence Encoder" can be found in figure~\ref{fig:sentence_encoder}.

    The scoring function takes one CRE vector and two entity embedding vectors as inputs, and outputs a scalar score.
\end{figure*}

\citet{weston-2013} conducted the first research work that utilized knowledge bases as a structural graph by training a TransE (\citet{transe}) KB model alongside a traditional distant supervision extractor. However, this has been a simple combination of both ideas but despite its simplicity, the model was able to beat the then-state-of-the-art models in Relation Extraction tasks. 
The work by \citet{han-2018}, on the other hand, went a step further by sharing some model weights between the distant supervision extractor and the knowledge base model. However, although there is some shared architecture, but the objective function for the KB modelling and the distant supervision remain separate. Another notable work by \citet{xu-2019} combines the two models through an additional objective function that guides the training but during prediction time there is no any shared architecture and hence knowledge base information is not really incorporated well for relation prediction.

The model presented in this paper differs from all previous work in that it is a single relation extraction model that internalizes knowledge base modeling. Instead of treating relation embeddings as parameters like in standalone knowledge embedding models (eg: TransE \cite{weston-2013}), where such embeddings have no context of any textual information, our model expresses relation embeddings as context-aware latent states generated by encoding textual data. Such contextualized embeddings represent a natural link between the textual input and the knowledge base modeling objective, so that the end-to-end information transformation is completely internal to the model. The entity embeddings, on the other hand, are not contextualized and will still be trained as parameters of the same model, because they need to serve as bridges across different entity pairs and globalize the knowledge base modeling which is now based on contextualized relation embeddings localized to each entity pair. A significant innovation of this work is that each sentence in which a pair of entities occurs, we represent each possible relation between them as a function of that entire sentence and score the relation for that entity pair and sentence using the representation of the relation. 

\section{Contextualized Relation Embedding}

Let $\mathcal{R}$ stand for the set of all relations and $\mathcal{E}$ stand for set of all entities. Then, for the $i$-th entity pair, $H_i$ 
and $T_i$ stand for the subject and object respectively, both belonging to $\mathcal{\mathcal{E}}$. $\mathcal{R}^{\prime}_i \subset \mathcal{R}$ 
stand for the set of relations existing between the $i$-th entity pair. $S_i$ stands for the collection of sentences that 
contain $H_i$ as the subject and $T_i$ as the object; $S_i^{jk}$ stands for the $k$-th word of the $j$-th sentence in $S_i$, 
where each $S_i^{jk}$ is delimited by spaces except for the subject and the object, whose phrasal integrities are preserved 
(e.g. if it is the subject/object of a sentence, \emph{the United States} will be considered as a single ``word'').

\begin{figure*}
    \includegraphics[width=\textwidth]{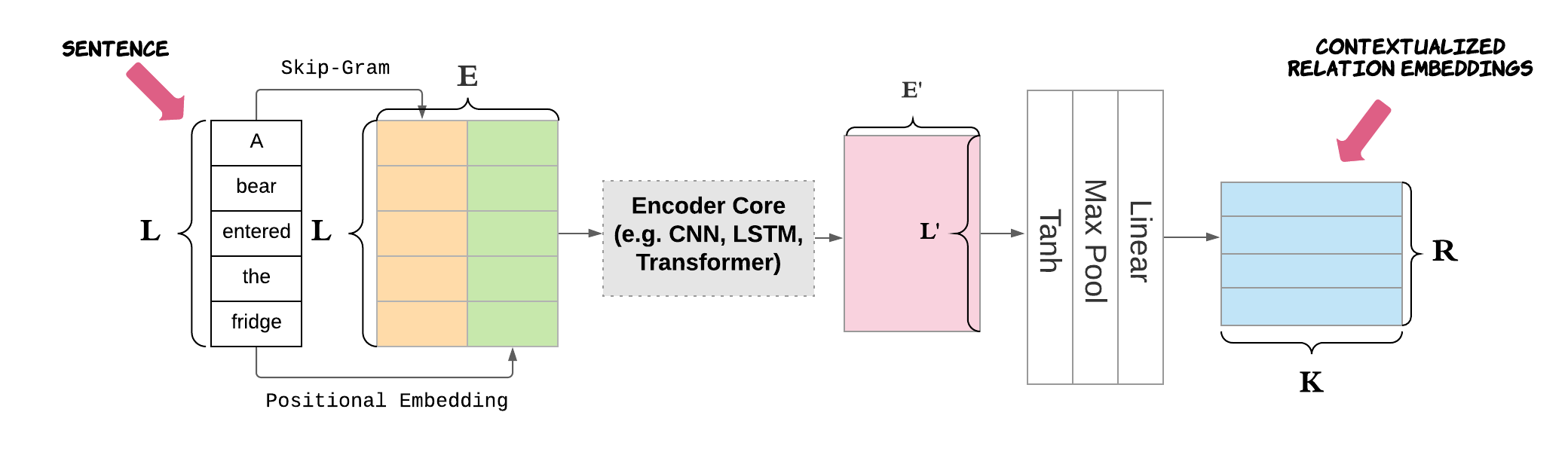}
    \caption{Single Sentence Encoding Process}\label{fig:sentence_encoder}
\end{figure*}

We propose a novel neural network model (called CRE model) that implicitly combines knowledge base to detect relationships between entities, given a corpus of sentences containing those entities. The overall architecture of a CRE model is shown in figure~\ref{fig:model}, which consists of a sentence encoding step and a knowledge base modeling step. During the sentence encoding step, the model transforms each $S_i^{jk}$ into $E$-dimensional 
embeddings. Specifically, we apply pretrained skip-gram embeddings \citet{word2vec}, treating all named entities as unknown, 
which results in a $W$-dimensional embedding for each word. We apply a positional embedding based on \citet{zeng2014}, where 
each word's index in the positional ``vocabulary'' is defined as its relative distance to one of the named entities. Consider the sentence 
``A bear entered the fridge'' and named entities ``bear'' and ``fridge''. The word ``A'' will have $-1$ as its positional embedding index 
with respect to ``bear'' and $-4$ with respect to ``fridge'', whereas ``entered'' will have $+1$ with respect to ``bear'' and $-2$ with 
respect to ``fridge''. For each word (including the two named entities), we represent it as its skip-gram word embedding of dimension $W$, concatenated with its positional embeddings (each of dimension $P$) with respect to the two named entities, to form a final embedding of dimension $E$, where $E = W + 2 * P$. The skip-gram embeddings are fixed during training, whereas the positional embeddings are learnt. This leads to a sentence embedding of $L * E$, where $L$ is the maximum length of each sentence. We utilize the above mentioned sentence embeddings and learn an encoder that encodes each sentence into a $R * K$ embedding, which can be represented as $|\mathcal{R}|$ $K$-dimensional embeddings, each of which corresponds to a distinct relation in $\mathcal{R}$. Figure~\ref{fig:sentence_encoder} illustrates the sentence encoding process for a single sentence.

It is worth noting that the sentence encoder is a generic framework where
we can plug in any model architecture (both for positional encoding and sentence encoding). In our experiments, we found that neural network based encoders tend to result in better performance. We think this is due to their ability to extract both semantic and syntactic information. We denote sentence $j$'s 
representation of $r$'s relation for entity pair $i$ as $CRE_{i}^{jr}$.

In order to provide the model more context, we explicitly apply a $K$-dimensional embedding for each part of the entity pair represented by $H_i$ and $T_i$ respectively. These entity embeddings are learnt as parameters during training, similar to the positional embeddings above. We utilize the relation embeddings (contextualized relation embeddings) from the sentences and these entity embeddings to form relation triplet embeddings. Specifically, this will result in $R$ triplets for each sentence, since we computed $R$ relation embeddings for each sentence. This allows us to apply different knowledge based models and further enhances the generalization of our model. Depending on the choice of knowledge based model, we use the relation triplet embeddings to compute its score based on the scoring function of the respective knowledge base model. For example, if TransE \citet{transe} is chosen as the underlying knowledge base model with $H_i$ and $T_i$ as head and tail entity embeddings, then embeddings for relation $r$ derived from sentence $j$ will be scored as:
\begin{equation}
    \begin{split}
        Score_{i}^{j, r} = 1 - tanh(||H_i
        + CRE_{i}^{jr} - T_i||) ; \\
        where\ 0\leq Score_{i}^{j, r}\leq 2
    \end{split}
\end{equation}

Since most knowledge base models are characterized by their scoring functions, it is straightforward to swap in any knowledge base model 
here for scoring purposes.

Finally, the model aggregates scores for the same relation across all sentences so that we can obtain a single score for each relation given an entity pair:

\begin{equation}
    \begin{split}
        Score_{i}^{r} = \sum_j Score_{i}^{j, r}
    \end{split}
\end{equation}

For experiments in this paper, we adopted summation as the aggregation function since we do not have prior knowledge about our dataset and 
thus want to take into account the ``opinion'' of all sentences equally, but it is by no means prescriptive. One can certainly 
use maximum, minimum or mean as the aggregation functions instead of sum in case the dataset is such that any single sentence's outcome can be regarded as reliable.

We train the model with a binary cross entropy loss, after normalizing relation scores and targets and we use an $L2$ regularization that helps prevent overfitting.

\begin{equation}
    \begin{split}
        NS_{i}^{r} = \frac{Score_{i}^{r}}{\sum_{r^{\prime}}Score_{i}^{r^{\prime}}}
    \end{split}
\end{equation}

\begin{equation}
    \begin{split}
        m_{i}^{r} = \frac{I[r \in \mathcal{R}^{\prime}_i]}{|\mathcal{R}^{\prime}_i|}
    \end{split}
\end{equation}

\begin{equation}
    \begin{split}
        L_{i}^{r} = -m_{i}^{r} * log(NS_{i}^r)) + \\
        (m_{i}^{r} - \frac{1}{|\mathcal{R}^{\prime}_i|}) * log(1 - NS_{i}^r) \\
    \end{split}
\end{equation}

\begin{equation}
    L_i = \sum_rL_{ir} + \lambda * \sum ||w||^2
\end{equation}

We use the normalized scores to predict the relations for the given entity pair. Note that normalization is still necessary for quantitative evaluation. We pool together scores for different entity pairs to form a precision-recall curve, so scores need to be comparable and normalization brings them to the same scale. The top-k prediction can formulated as:
\begin{equation}
    \begin{split}
        Top_{i}^{k} = argmax_{r}^{k}(NS_{i}^{r}) \\
        TopScore_{i}^{k} = NS_{i}^{Top_{i}^{k}}
    \end{split}
\end{equation}

\section{Experiments}
\subsection{Data and Experimental Setup}
For our experiments, we use the textual data from The New York Times (NYT) Annotated Corpus \citet{riedel2010}, and the knowledge base are derived from the most 
recent FreeBase \citet{freebase} dump. Note that entities in each NYT sentence are already annotated. The NYT and Freebase presents a more challenging task for relation extraction models than FreeBase and Wikipedia texts due to their heterogeneous nature, since FreeBase itself is largely derived from Wikipedia. We construct the dataset through the following procedure:

\begin{enumerate}
    \item Filter out any relation instance from FreeBase dump if the mapping between the ID of either of its two entities and its corresponding English phrase is not available.
    \item Find the top 500K entities based on the number of relation instances they participate in, and further filter out any relation instance if either of its two entities is not in the top 500K list.
    \item Inner join the NYT dataset with the filtered Freebase by aligning NYT entity annotations and Freebase entity English phrases. In other words, no example in the NYT dataset that contains unseen entities in the filter Freebase is preserved, and vice versa.
    \item Backfill an N/A relation for any entity pair present in the filtered NYT dataset that has no relation instance in the filtered Freebase.
\end{enumerate}

After this procedure, we have a dataset that contains 465K sentences, 35K entities, 233K entity pairs, and 238 relations. As most entities 
in this world do not possess direct relations with other entities, the dataset is extremely unbalanced, where only 5K entity pairs out of the 233K have non-N/A relation. These 5K entity pairs contribute 20K sentences out of the 465K. Because of this imbalance, we performed train-test splits for positive pairs and negative pairs separately, so that test set gets around 1200 positive pairs and 57K negative pairs. Since this test set is still severely imbalanced and we don’t want our model to be biased in favor an N/A prediction, we further randomly sample the negative pairs in the test set so that the number of sentences consisting of negative pairs is roughly the same as that of any positive relation. We did not perform similar filtering to the training set from the outset, but randomly applied such filtering at the beginning of each training epoch, in order to utilize 
all training data we have.

\begin{figure*}
    \includegraphics[width=0.9\textwidth]{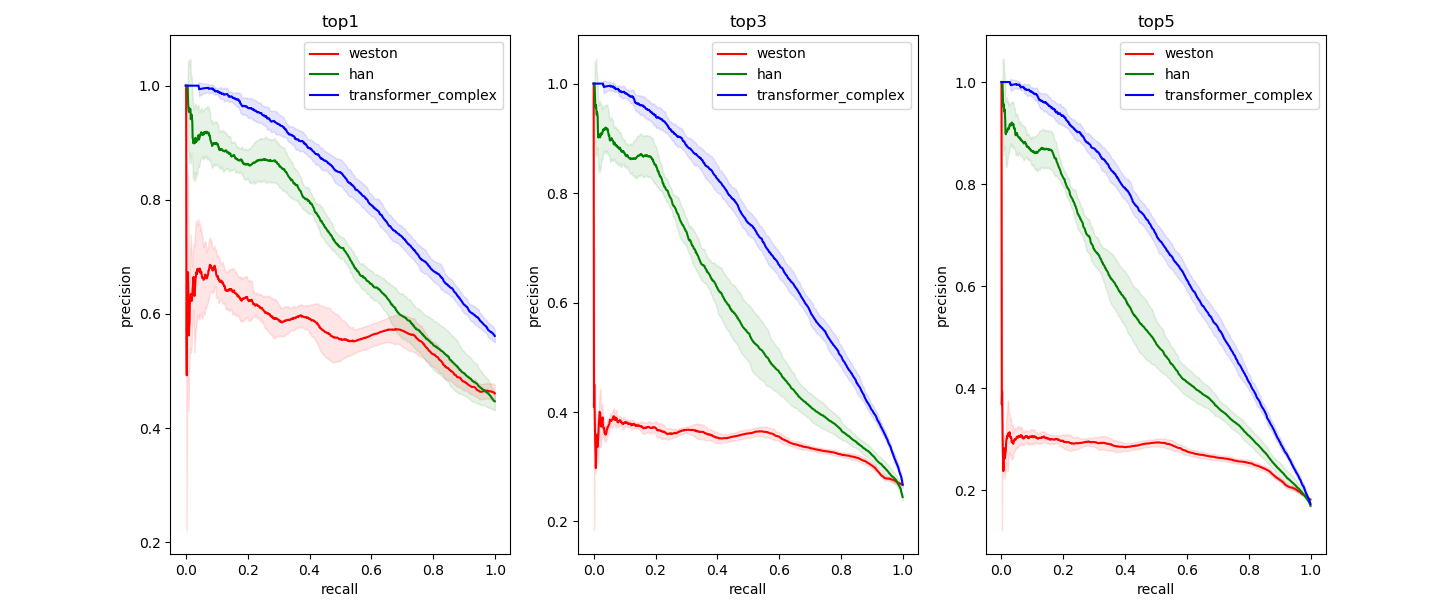}
    \caption{Weston et al. vs. Han et al. vs. CRE Model}\label{fig:comparison}
\end{figure*}

\subsection{Evaluation Metrics}
In order to quantitatively evaluate the performance of our model, we measure precision over various recall levels. Specifically, we examine precision-recall curves for top-1 predictions, top-3 predictions, and top-5 predictions. Top-1 predictions, which only includes the relation with the highest score among all relations for a given entity pair, are widely used in related literature, but we believe it alone does not provide a comprehensive view, since each entity pair can have multiple relations. Besides the precision-recall curve, we also measure the mean reciprocal rank (MRR) among top-3 predictions and top-5 predictions.

On the qualitative side, we examine the number of distinct top-1 relation predictions, in order to check if our model achieves high precision just by heavily favoring certain relations, given the unbalanced nature of our dataset.

\subsection{Baseline Models}
As discussed in Section 2, our proposed model provides a deep integration of knowledge bases in the relation prediction task. 
We select two baseline models to compare performance: one from \citet{weston-2013} 
as a representative of weak integration between knowledge bases and relation extraction, and the other from \citet{han-2018} to represent a semi-integrated setup. Both models have already been briefly described in the Introduction section. Since the state of art of these models is not publically available, we performed hyperparameter tuning for both models on our dataset, since they may not be exactly the same as the datasets used in their respective work.

\subsection{Model Configurations}
As shown in the model details section, there can be a myriad of options for various components of our proposed CRE model. In this evaluation, we will focus on the effect of different choices of sentence-to-CRE encoders as well as underlying knowledge base models. Every other aspect will be kept fixed as described in the preceding section.

For sentence-to-CRE encoders, we will explore 3 options:
\begin{itemize}
    \item A single 1D convolutional layer with hidden state dimension of 230 and window size of 3.
    \item A single LSTM layer with hidden state dimension of 230.
    \item A double-layered transformer encoder with hidden state dimension of 100 and 5-head attention \citet{Vaswani}.
\end{itemize}
All these encoders are followed by a $tanh$ activation layer and a linear layer to project the low-dimensional hidden state onto $|\mathcal{R}|K$-dimensional 
space. 230 is selected as the hidden dimension for fair comparison purpose because it is the CNN hidden state dimension used in \citet{han-2018}. The 
complexity of encoders evaluated here are limited by computational resources available for this work, so it is possible to find more sophisticated encoders that can achieve further improvements. For the transformer based encoder, an additional linear projection and tanh activation is applied at the front to 
reduce the number of parameters that need to be trained to a manageable level. 
%Due to computational resource constraints, we will not perform cross-validation to select hyperparameters except learning rate and L2 regularization coefficients.

\begin{figure*}
    \includegraphics[width=\textwidth]{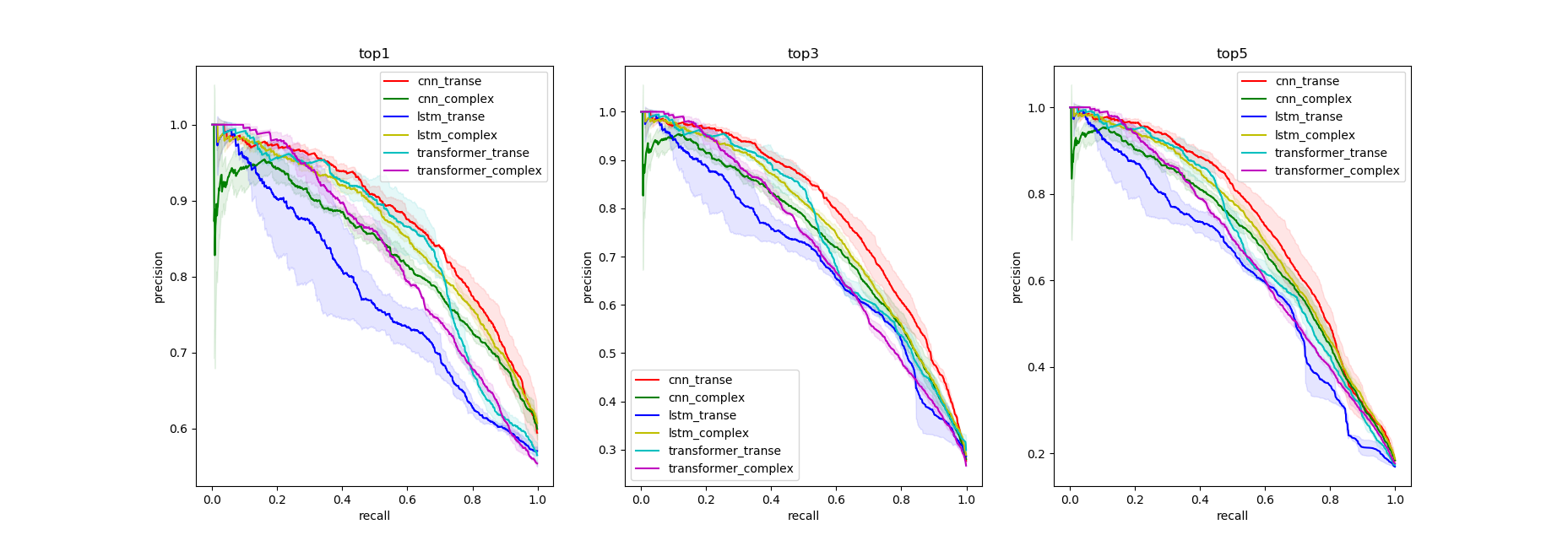}
    \caption{CRE Models with different configurations}\label{fig:comparison_2}
    \medskip
    \small
    Left: precision-recall curves of the most confident prediction for each entity pairs

    Middle: precision-recall curves of the top 3 most confident predictions for each entity pairs

    Right: precision-recall curves of the top 5 most confident predictions for each entity pairs
\end{figure*}

\begin{table*}[t]
    \begin{tabularx}{\textwidth}{X||l|l}
        \textbf{Model} & \textbf{MRR among Top-3} & \textbf{MRR among Top-5} \\
        \hline
        Weston et al. & 0.6010 & 0.6194 \\
        Han et al. & 0.5786 & 0.6005 \\
        CRE(transformer+ComplEx) & \textbf{0.6413} & \textbf{0.6588} \\
        \hline
    \end{tabularx}
\caption{\label{tab:mrr-baseline}MRR comparison with baseline models}
\end{table*}
\begin{table*}[t]
    \begin{tabularx}{\textwidth}{X||l|l}
        \textbf{Model} & \textbf{MRR among Top-3} & \textbf{MRR among Top-5} \\
        \hline
        CNN+TransE & \textbf{0.6925} & \textbf{0.7059} \\
        CNN+ComplEx & 0.6681 & 0.6805 \\
        LSTM+TransE & 0.6234 & 0.6506 \\
        LSTM+ComplEx & 0.6802 & 0.6938 \\
        transformer+TransE & 0.6332 & 0.6549 \\
        transformer+ComplEx & 0.6413 & 0.6588 \\
        \hline
    \end{tabularx}
\caption{\label{tab:mrr-config}MRR comparison between different configurations of CRE}
\end{table*}

For knowledge base model, we explore 2 options:
\begin{itemize}
    \item TransE with embedding dimension of 50.
    \item ComplEx \citet{complex} with embedding dimension of 25 for the real part and 25 for the imaginary part.
\end{itemize}
The dimensions chosen here are based on reports from each model’s respective paper. 

%The comment regarding model complexity and cross-validation above also applies here.

By combining these options, we will explore 6 different model configurations in total. We use Adam optimizer for training and consider the model to have converged when the loss on the current epoch is no less than the average loss of the last ten epochs. Each model is then used to make predictions on the same held-out test set.

In order to test for statistical significance, we divided the dataset into 3 different subsets randomly and recorded the results of each model on all three subsets.

\subsection{Results}
\subsubsection{The best model configuration}
We compare the performance of the CRE models across different configurations, which is illustrated in figure~\ref{fig:comparison_2}. From the results, we can see that the more sophisticated knowledge base model (ComplEx) outperforms the simpler alternative (TransE) at lower recall levels (less than 0.25), but under-performs at higher recall levels. Similarly, for different choices of sentence-to-CRE encoders, we observe that the more complex transformer encoder is almost perfect before recall rises above 0.25. 

The most consistent configuration, from a quantitative perspective, is the combination of CNN and TransE. These results are reinforced by
the MRR comparison among different CRE configurations, as shown in table~\ref{tab:mrr-config}. We can see that the model that achieved the best precision-recall result (CNN+TransE) also dominates in terms of MRR, whereas the worst results for both metrics are attributable to the LSTM+TransE configuration.

\subsubsection{Comparison with baseline models}

Figure~\ref{fig:comparison} shows the precision-recall curve comparisons between baseline models and a CRE-based model with Transformer+ComplEx configuration. Each model's  
confidence interval was obtained via variance among precisions at the same recall level of 9 different runs: 3 runs for each of 3 different random subset of the training/testing dataset. The CRE model was able to outperform both baseline models. 

Interestingly, Weston’s approach achieved top-1 prediction accuracy comparable with what was reported in its original paper, but saw a particularly sharp drop in precision as the number of predictions examined increases. This contrast is unsurprising though, because Weston's approach only re-scores its most confident predictions, so models trained this way have no capability of making multiple predictions for a single entity pair.

In addition, we observed the mean reciprocal rank for CRE model is significantly higher than baseline models, as shown in table~\ref{tab:mrr-baseline}, which corresponds to the improvement we saw in terms of precision-recall curve.

\subsubsection{Qualitative evaluation}
Table~\ref{tab:relation-counts} shows most frequent relations and the number of times each model concludes one of them to be the most likely relation given an entity pair, as well as the ground truth. It can be observed that the CRE models tend to generate a less skewed distribution of these frequent relations compared to Han’s, which is the better of the two baseline models in terms of quantitative performance. 

\begin{table*}[t]
    \begin{tabularx}{\textwidth}{X||l|l|l|l|l}
        \textbf{Relation Name} & \textbf{Weston et al.} & \textbf{Han et al.} & \textbf{CRE+CNN} & \textbf{CRE+Transformer} & \textbf{Fact} \\
        \hline
        location.contains & 1085 & 1178 & 761 & 840 & 597 \\
        person.nationality & 66 & 19 & 189 & 159 & 173 \\
        location.containedBy & 198 & 205 & 435 & 346 & 316 \\
        people.placeOfBirth & 7 & 0 & 16 & 17 & 91 \\
        people.placeOfDeath & 1 & 0 & 0 & 22 & 47 \\
        usRepresentative.state & 0 & 0 & 0 & 5 & 11 \\
        tvProgram.programCreator & 0 & 0 & 0 & 1 & 2 \\
        \hline
    \end{tabularx}
\caption{\label{tab:relation-counts}Top relation prediction counts by each model compared to truth}
\end{table*}

\section{Discussion}
Our experiments from figure \ref{fig:comparison} clearly demonstrate that models utilizing contextualized relation embeddings that internalize both relation 
extractor modeling and knowledge base modeling tend to perform much better on relation extraction task than architectures like \citet{weston-2013} and \citet{han-2018} that join a relation extractor and a knowledge base model in some arbitrary way. We believe this is due to the fact that the ``internal knowledge base models'' within CRE models are context-aware. As a result, CRE models can take advantage of the contextual information contained in the corpus more effectively.

We demonstrate the generality and flexibility of the proposed CRE model that can work with different encoders and knowledge base models. Moreover, it can be observed from Table \ref{tab:relation-counts} that the CRE-models generate less skewed distribution of frequent relations compared to the baseline models, thus demonstrating that the CRE model provides robust predictions and works well even with imbalanced datasets. 

It can be observed from Figure \ref{fig:comparison_2} that more complex configurations for our model tend to achieve stellar results in the low recall arena but lose steam when recall levels are high. This may be explained by the insufficient training epochs due to resource constraints. As a result, these complex models did not get the chance to optimize for the less seen prediction targets. However, it can been seen from Table \ref{tab:relation-counts} that the transformer-based CRE model was able to correctly uncover some rare relations, like \emph{programCreator}, which was missed by the CNN-based CRE model, despite it having better overall quantitative results. It is reasonable to expect that given a more balanced dataset with sufficient training time, the transformer-based CRE model may obtain a better quantitative result than its CNN counterpart.

Since contextual knowledge plays a big role in the performance of CRE models, it will be interesting to see how such models may perform over prediction on input texts that are heterogeneous to the New York Times. For example, an alternative input text source could be a collection of academic publications, which is of very different genre compared to the New York Times. If our hypothesis on why these models work well is correct, then we would expect to see some degradation in performance, though it should be no worse than context-free approaches. This can be overcome, of course, by utilizing pre-trained CRE models and finetuning on the new text body, so that the model can learn about the new context.

\section{Conclusion and Future Work}
In this paper, we introduced a novel contextualized relation embedding (CRE) model for relation extraction that incorporates knowledge base modeling in a comprehensive and efficient manner. We demonstrate both empirically and qualitatively that the CRE model is able to achieve state of the art results on relation extraction task on the New York times dataset with Freebase as a knowledge source. We demonstrate the flexibility of the model by using different encoders and knowledge based modeling schemes which are a testament to the modular nature of this model, which can easily be upgraded with alternative configurations with ease. Finally, we showcase that CRE model tend to generate a less skewed distributions of predicted relations and the model is robust to imbalances in the dataset.

Interesting directions of future work may include utilizing CRE pre trained models and evalauting if these models can be finetuned to work in low-resource situations when the testing set distribution is different than the training distribution. Another interesting direction may be utilizing pre-trained BERT-based \citet{devlin2018bert} models as sentence encoders and fine-tune them for the relation extraction task, since such models are already trained on large amount of textual data and thus may hold contextual knowledge that might be useful for this task. This may also reveal some hidden connections between contextualized word embeddings and contextualized relation embeddings.

\bibliography{emnlp2020}
\bibliographystyle{acl_natbib}

\appendix
\end{document}